\documentclass{clv3}

\usepackage{hyperref}
\usepackage{xcolor}
\usepackage{makecell}
\definecolor{darkblue}{rgb}{0, 0, 0.5}
\hypersetup{colorlinks=true,citecolor=darkblue, linkcolor=darkblue, urlcolor=darkblue}

\usepackage{multirow}

\usepackage{amsmath}

\usepackage{tabularx}
\newcolumntype{R}[1]{>{\hsize=#1\hsize\raggedleft\arraybackslash}X}%
\newcolumntype{L}[1]{>{\hsize=#1\hsize\raggedright\arraybackslash}X}%
\newcolumntype{C}[1]{>{\hsize=#1\hsize\centering\arraybackslash}X}%

\historydates{Submission received: 28 April 2020; revised version received: 12 November 2020; accepted for publication: 8 December 2020.}

\issue{47}{1}{2021}

\runningtitle{Interpretability Analysis for NER}

\runningauthor{Agarwal, Yang, Wallace and Nenkova}

\begin{document}

\title{Interpretability Analysis for Named Entity Recognition to Understand System Predictions and How They Can Improve}

\author{Oshin Agarwal}
\affil{University of Pennsylvania \\ Department of Computer and Information Science \\ \texttt{oagarwal@seas.upenn.edu}}

\author{Yinfei Yang}
\affil{Google Research \\ \texttt{yinfeiy@google.com}}

\author{Byron C. Wallace}
\affil{Northeastern University \\ Khoury College of Computer Sciences \\ \texttt{b.wallace@northeastern.edu}}

\author{Ani Nenkova}
\affil{University of Pennsylvania \\ Department of Computer and Information Science \\ \texttt{nenkova@seas.upenn.edu}}

\maketitle

\begin{abstract}
Named entity recognition systems achieve remarkable performance on domains such as English news. It is natural to ask: What are these models actually learning to achieve this? Are they merely memorizing the names themselves? Or are they capable of interpreting the text and inferring the correct entity type from the linguistic context? We examine these questions by contrasting the performance of several variants of architectures for named entity recognition, with some provided only representations of the context as features. We experiment with Glove-based BiLSTM-CRF as well as BERT. We find that context does influence predictions, but the main factor driving high performance is learning the name tokens themselves. Furthermore, we find that BERT is not always better at recognizing predictive contexts compared to a BiLSTM-CRF model. We enlist human annotators to evaluate the feasibility of inferring entity types from context alone and find that humans are also mostly unable to infer entity types for the majority of examples on which the context-only system made errors. However, there is room for improvement: A system should be able to recognize any named entity in a predictive context correctly and our experiments indicate that current systems may be improved by such capability. Our human study also revealed that systems and humans do not always learn the same contextual clues, and context-only systems are sometimes correct even when humans fail to recognize the entity type from the context. Finally, we find that one issue contributing to model errors is the use of `entangled' representations that encode both contextual and local token information into a single vector, which can obscure clues. Our results suggest that designing models that explicitly operate over representations of local inputs and context, respectively, may in some cases improve performance. In light of these and related findings, we highlight directions for future work.
\end{abstract}

\section{Introduction}

Named Entity Recognition (NER) is the task of identifying words and phrases in text that refer to  
a person, location or organization name, or some finer subcategory of these types. NER systems work well on domains such as English news, achieving high performance on standard datasets like MUC-6 \cite{grishman1996message}, CoNLL 2003 \cite{tjong-kim-sang-de-meulder-2003-introduction} and OntoNotes \cite{pradhan-xue-2009-ontonotes}. However, prior work has shown that the performance deteriorates on entities unseen in the training data \cite{augenstein2017generalisation,fu2020rethinking} and when entities are switched with a diverse set of entities even within the same dataset \cite{agarwal2020entityswitched}.

In this paper, we examine the interpretability of models used for the task, focusing on the type of textual clues that lead systems to make predictions. Consider, for instance, the sentence  ``Nicholas Romanov abdicated the throne in 1917''. 
The correct identification of ``Nicholas Romanov'' as a person may be due to {\em (i)} knowing that {\em Nicholas} is a fairly common name and that {\em (ii)} the capitalized word after that ending with "-ov" is likely a Slavic last name too. Alternatively, {\em (iii)} a competent user of language would know the selectional restrictions \cite{framis-1994-experiment,akbik-etal-2013-effective,chersoni-etal-2018-modeling} for the subject of the verb abdicate, i.e., that only a person may abdicate the throne. The presence of two words indicates that it cannot be a pronoun, so $X$ in the context ``$X$ abdicated the throne" can only be a person. 

Such probing of the reasons behind a prediction is in line with early work on NER that emphasized the need to consider both internal (features of the name itself) and external (context features) evidence when determining the semantic types of named entities \cite{mcdonald1993internal}. We specifically focus on the interplay between learning names as in {\em (i)}, and recognizing constraining contexts as in {\em (iii)}, given that {\em (ii)} can be construed as a more general case of {\em (i)}, in which word shape and morphological features may indicate that a word is a name even if the exact name is never explicitly seen by the system (Table 1 in \cite{bikel1999algorithm}). 

Below are some examples of constraining contexts for different entity types. The type of X in each of these contexts should always be the same, irrespective of the identity of X.

\begin{description}
\item [PER] My name is $X$.
\item [LOC] The flight to $X$ leaves in two hours.
\item [ORG] The CEO of $X$ resigned.
\end{description}

As a foundation for our work, we conduct experiments with BiLSTM-CRF models using GloVe input representations \cite{huang2015bidirectional} modified to use \emph{only context} representations or \emph{only word} identities to quantify the extent to which systems exploit word and context evidence, respectively (Section \ref{sec:analysis}). We test these systems on three different datasets to identify trends that generalize across corpora. 
We show that context does somewhat inform system predictions, but the major driver of performance is recognition of certain words as names of a particular type. We modify the model by introducing gates for word and context representations to determine what it focuses on. We find that on average, only the gate value of the word representation changes when there is a misprediction; the context gate value remains the same. We also examine the performance of a BERT-based NER model and find that it does not always incorporate context better than the BiLSTM-CRF models.

We then ask if systems should be expected to do better from the context text (Section \ref{sec:humaneval}). Specifically, we task human annotators with inferring entity types using only (sentential) context, for the instances on which a BiLSTM-CRF relying solely on context made a mistake. We find that in the majority of cases, annotators are (also) unable to choose the correct type. This suggests that it may be beneficial for systems to similarly recognize situations in which there is a lack of reliable semantic constraints for determining the entity type. annotators sometimes make the same mistakes as the model does. This may hint at why conventional systems tend to ignore context features: The number of examples for which relying primarily on contextual features will result in an accurate prediction is almost the same as the number for which relying on the context will lead to an erroneous prediction. For some cases, however, annotators are able to correctly identify the entity type from the context alone when the system makes an error, indicating that systems do not identify all constraining contexts and that there is room for improvement.

We conclude by running `oracle' experiments in Section \ref{sec:analysis} that show that systems with access to different parts of the input can be better combined to improve results. These experiments also show that the Glove-based BiLSTM-CRF and BERT based only on the context representation are correct on very different examples; consequently, combining the two leads to performance gains. There is thus room for better exploiting the context to develop robust NER systems with improved generalizability. 

Finally, we discuss the implications of our findings for the direction of future research, summarizing the answers to three questions we explore in this paper - 1) What do systems learn - word identity vs context?, 2) Can context be utilized better so that generalization is improved?, 3) If so, what are some possible directions for it? .

\section{Related Work}

Most related prior work has focused on learning to recognize certain words as names, either using the training data, gazetteers or, most recently, pre-trained word representations. Early work on NER did explicitly deal with the task of scoring contexts on their ability to predict the entity types in that context. More recent neural approaches have only indirectly incorporated the learning of context, namely via contextualized word representations \cite{peters-etal-2018-deep,devlin-etal-2019-bert}.

\subsection{Names Seen in Training} 

NER systems recognize entities seen in training more accurately than entities that were not present in the training set \cite{augenstein2017generalisation, fu2020rethinking}.
The original CoNLL NER task used as a baseline a name look-up table: Each word that was part of a name that appeared in the training data with a unique class was correspondingly classified in the test data as well. All other words were marked as non-entities. Even the simplest learning systems outperform such a baseline \cite{tjong-kim-sang-de-meulder-2003-introduction}, as it will clearly achieve poor recall. At the same time, overviews of NER systems indicate that the most successful systems, both old \cite{tjong-kim-sang-de-meulder-2003-introduction} and recent \cite{yadav2018survey}, make use of gazetteers listing numerous names of a given type. Wikipedia in particular has been used extensively as a source for lists of names of given types \cite{kazama2007exploiting,ratinov2009design}. The extent to which learning systems are effectively `better` look-up models --- or if they actually learn to recognize contexts that suggest specific entity types --- is not obvious.

Even contemporary systems that do not use gazetteers expand their knowledge of names through the use of pre-trained word representations.  
With distributed representations trained on large background corpora, a name is ``seen in training" if its representation is similar to names that explicitly appeared in the training data for NER. Consider, for example, the commonly used Brown cluster features \cite{brown1992class,miller2004name}. Both in the original paper and the re-implementation for NER, authors show examples of representations that would be the same for classes of words (John, George, James, Bob or John, Gerald, Phillip, Harold, respectively). In this case, if one of these names is seen in training, any of the other names are also treated as seen, because they have the exact same representation. 

Similarly, using neural embeddings, words with representations similar to those seen explicitly in training would likely be treated as ``seen" by the system as well. Tables 6 and 7 in \cite{collobert2011natural} show the impact of word representations trained on small training data also annotated with entity types compared to those making use of large amounts of unlabeled text. When using only the limited data, the words with representations closest to {\em france} and {\em jesus} respectively are ``persuade, faw, blackstock, giorgi" and ``thickets, savary, sympathetic, jfk", which seem unlikely to be useful for the task of NER. For the word representations trained on Wikipedia and Reuters,\footnote{CoNLL data, one of the standard datasets used to evaluate NER systems, is drawn from Reuters.} the corresponding most similar words are ``austria, belgium, germany, italy" and ``god, sati, christ, satan". These representations clearly have higher potential for improving NER. 

Systems with character-level representations further expand their ability to recognize names via word shape (capitalization, dashes, apostrophes) and basic morphology \cite{lample-etal-2016-neural}.

We directly compare the a lookup baseline with a system that uses only predictive contexts learned from the training data, and an expanded baseline drawing on pre-trained word representations which cover many more names than the limited training data itself.  

\subsection{Unsupervised Name-Context Learning} 

Approaches for database completion and information extraction use free unannotated text to learn patterns predictive of entity types \cite{riloff1999learning,collins1999unsupervised,agichtein2000snowball,etzioni2005unsupervised,banko2007open} and then use these to find instances of new names. Given a set of known names, they rank all $n$-gram contexts for their ability to predict the type of entities, discovering for example that ``the mayor of X" or ``Mr. Y" or ``permanent resident of Z" are predictive of city, person, and country respectively. 

Early NER systems also attempted to use additional unannotated data, mostly to extract names not seen in training but also to identify predictive contexts. These however had little to no effect on system performance \cite{tjong-kim-sang-de-meulder-2003-introduction}
with few exceptions where both names and contexts were bootstrapped to train a system \cite{cucerzan2002language,nadeau2006unsupervised,talukdar2006context}. 

Recent work in NLP relies on neural representations to expand the ability of the systems to learn context and names \cite{huang2015bidirectional}. In these approaches the learning of names is powered by the pre-trained word representations, as described in the previous section, and the context is handled by an LSTM representation. So far, there has not been analysis of which parts of contexts are properly captured by the LSTM representations, especially what they do better than more local representations of just the preceding/following word.

Acknowledged state-of-the-art approaches have demonstrated the value of \emph{contextualized} word embeddings, as in ELMo~\cite{peters-etal-2018-deep} and BERT~\cite{devlin-etal-2019-bert}; these are representations derived both from input tokens and the context in which they appear. They have the clear benefit of making use of large datasets for pre-training that can better capture a diversity of contexts. But at the same time these contextualized representations make it difficult to interpret the system prediction and which parts of the input led to a particular output. Contextualized representations can in principle disambiguate the meaning of words based on their context, e.g., the canonical example of Washington being a person, a state or a city, depending on the context. This disambiguation may improve the performance of NER systems. Furthermore, token representations in such models reflect their context by construction, so may specifically improve performance on entity tokens not seen during training but encountered in contexts that constrain the type of the entity.
 
To understand the performance of NER systems, we should be able to probe the justification for the predictions: Did they recognize a context that strongly indicates that whatever follows is a name of a given type (as in ``Czech Vice-PM \_''), or did they recognize a word that is typically a name of a given type (``Jane''), or a combination of the two? In this paper, we present experiments designed to disentangle to the extent possible the contribution of the two sources of confidence in system predictions. We perform in-depth experiments on systems using non-contextualized word representations and a human/system comparison with systems that exploit contextualized representations.

\section{Context-only and Word-only Systems}
\label{sec:analysis}

Here we perform experiments to disentangle the performance of systems based on the word identity and the context. 
We compare two look-up baselines and several systems which vary the representations fed into a sequential Conditional Random Field (CRF) \cite{lafferty2001conditional}, described below.

\paragraph{Lookup} Create a table of each word preserving its case, and its most frequent tag from the training data. In testing, lookup a word in this table and assign its most frequent tag. If the word does not appear in the training data or there is a tie in the tag frequency, mark as O (outside, not a named entity).

\paragraph{LogReg} Logistic Regression using the GloVe representation of the word only (no context of any kind). This system is equivalent to lookup in both the NER training data and GloVe representations as determined by the data they were trained on. 

\paragraph{GloVe fixed + CRF} This system uses GloVe word representations as features in a linear chain CRF model. Any word in training or testing that does not have a GloVe representation is assigned representation equal to the average of all words represented in GloVe. The GloVe input vectors are fixed in this setting, i.e., we do not backpropagate into these. 

\paragraph{GloVe fine-tuned + CRF} The same as the preceding model, except that GloVe embedding parameters are updated during training. This method nudges word representations to become more similar depending on how they manifest in the NER training data, and generally performs better than relying on fixed representations.  

\paragraph{FW context + CRF} This system uses LSTM \cite{hochreiter1997long} representations only for the text preceding the current word (i.e., run forward from the start to this word), with GloVe as inputs. Here we take the hidden state of the previous word as the representation of the current word. This incorporates non-local information not available to the two previous systems, from the part of the sentence before the word. 

\paragraph{BW context + CRF} Backward context-only LSTM with GloVe as inputs. Here we reverse the sentence sequence and take the hidden state of the next word in the original sequence as the output representation of the current word.
  
\paragraph{BI context + CRF} Bidirectional context-only LSTM \cite{graves2005framewise} with GloVe as input. We concatenate the forward and backward context-only representations taking the hidden state as in the two systems above and not the hidden state of the current word.
    
\paragraph{BI context + word + CRF} Bidirectional LSTM as in \cite{huang2015bidirectional}. The feature representing the word is the hidden state of the LSTM after incorporating the current word; the backward and forward representations are concatenated.

We use 300 dimensional cased GloVe \cite{pennington2014glove} vectors trained on Common Crawl.\footnote{http://nlp.stanford.edu/data/glove.840B.300d.zip} The models are trained for 10 epochs using Stochastic Gradient Descent with a learning rate of 0.01 and weight decay of 1e-4. A dropout of 0.5 is applied on the embeddings and the hidden layer dimension used for the LSTM is 100.
We use the IO labeling scheme and evaluate the systems via micro-F1, at the token level. We use the word-based model for all the above variations, but believe a character-level model would yield similar results: Such models would differ only in how they construct the independent context and word representations that we consider. 

While the above systems would show how the model behaves when it has access to only specific information -- context or word -- they do not capture what the model would focus on with access to both types of information. For this reason, we build a gated system as follows.

\begin{figure}
  \includegraphics[width=\textwidth,trim={1.5cm 9cm 1.5cm 1.5cm},clip]{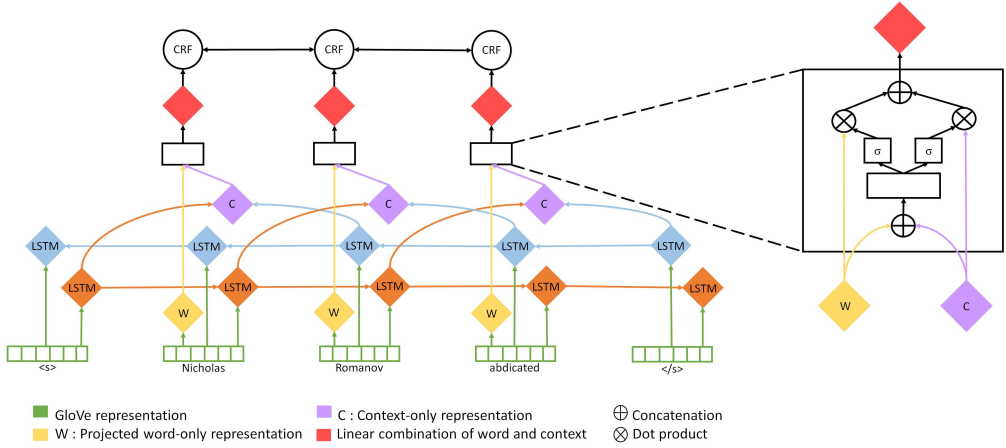}
  \caption{Architecture of the gated system. For each word, a token only (yellow) and a context only (purple) representation is learned. These are combined using gates, as illustrated on the right, and fed into a CRF.}
  \label{fig:gatedsystem}
\end{figure}

\paragraph{Gated BI + word + CRF} This system comprises a bidirectional LSTM that uses both context and the word, but the two representations are combined using gates. This provides a mechanism for revealing the degree to which the prediction was influenced by the word versus the context. 

More concretely, let $X = (x_1,x_2,...,x_n)$ be the input sequence where $x_i \in R^{d}$ is the pre-trained word embedding of the $i^{th}$ word in the sequence. The input representation of each word is projected to a lower dimensional space, yielding a word-only representation $x_i^w \in R^{d_w}$.

\begin{align*}
    x_i^w = W_w x_i
\end{align*}

For each word, we also induce a context-only representation using two LSTMs. One computes the representation of the forward context $h_i^{fw}$; the other reverses the sequence to derive a representation of the backward context $h_i^{bw}$. As noted earlier, these hidden states incorporate the representation of the $i^{th}$ word. Because we want a context-only representation, we instead use the hidden state for the previous word in the fw LSTM, and hidden state for the following word yielded from the bw LSTM. We concatenate these to form a context-only representation  $x_i^c \in R^{d_c}$.

\begin{align*}
    x_i^c = [h_{i-1}^{fw}; h_{i+1}^{bw}]
\end{align*}

\noindent Next we combine the word-only and context-only representations using parameterized gates:

\begin{align*}
    g_w = \sigma(W^{gw} [x_i^w; x_i^c]) \\
    g_c = \sigma(W^{gc} [x_i^w; x_i^c]) \\
    x_i^{wc} = g_w x_i^w + g_c x_i^c
\end{align*}

This is followed by a linear layer that consumes the resultant $x_i^{wc}$ representations and produces logits for potential class assignments for tokens that are subsequently decoded using a linear chain CRF.
This architecture is illustrated in Figure \ref{fig:gatedsystem}. 

In addition to the above systems that are based on GloVe representations, we also perform experiments using the following systems that use contextual representations, for the sake of completeness. 

\paragraph{Full BERT} We use the original public large model\footnote{\texttt{cased\_L-24\_H-1024\_A-16}} and apply the default fine-tuning strategy. We use the NER finetuning implementation in \cite{Wolf2019HuggingFacesTS} and train the model for 3 epochs with a learning rate of 5e-06 and maximum sequence length of 128. We use the default values of all other hyperparameters as in the implementation.

\paragraph{Word-only BERT} Fine-tuned using only the word as the input resulting in a non-contextual word-only representation. Each word token can appear multiple times with various entity types in the training data and the frequency for each is maintained in the training data.

\paragraph{Context-only BERT} Since decomposition of BERT representations into word-only and context-only is not straightforward,\footnote{We tried a few techniques such as projecting the BERT representations into word and context spaces by learning to predict the word itself and the context words as two simultaneous tasks, but this did not work well.} we adopt an alternate strategy to test how BERT fairs without seeing the word itself. We use a reserved token from the vocabulary `[unused0]' as a mask for the input token so that the system is forced to make decisions based on the context and does not have a prior entity type bias associated with the mask. We do this for the entire dataset, masking one word at a time. It is important to note that the word is only masked during testing and not during fine-tuning.

\begin{table}
\centering
\small
\setlength{\tabcolsep}{3pt}
\setlength{\extrarowheight}{1pt}
\caption{Performance of GloVe word-level BiLSTM-CRF and BERT. All rows are for the former and only the last two rows for BERT. Local context refers to high precision constraints due to sequential CRF. Non-local context refers to the entire sentence. No document level context is included. The first two panels were trained on the Original English CoNLL 03 training data and tested on the original English CoNLL 03 test data and the WikiGold data. The last panel was trained and tested on the respective splits of MUC-6. Highest F1 in each panel is boldfaced, excluding the full systems.}
\begin{tabularx}{\linewidth}{L{0.8}|C{0.2}C{0.2}C{0.2}|C{0.2}C{0.2}C{0.2}|C{0.2}C{0.2}C{0.2}}
\hline
\multirow{2}{*}{\textbf{System}}& \multicolumn{3}{c|}{\textbf{CoNLL}} & \multicolumn{3}{c|}{\textbf{Wikipedia}}& \multicolumn{3}{c}{\textbf{MUC-6}} \\
\cline{2-10}
&P&R&F1&P&R&F1&P&R&F1\\\hline
{\em Full system} & & & & & & & & & \\
BI context + word + CRF & 90.7 & 91.3 & 91.0 & 66.6 & 60.8 & 63.6 & 90.1 & 91.8 & 90.9\\  \hline \hline
{\em Words only} & & & & & & & & & \\
Lookup & 84.1 & 56.6 & 67.7 & 66.3 & 28.5 & 39.8 & 81.4 & 54.4 & 65.2 \\
LogReg & 80.2 & 74.3 & \textbf{77.2} & 58.8 & 48.9 & \textbf{53.4} & 75.1 & 71.7 & \textbf{73.4}\\ \hline \hline
{\em Words + local context} & & & & & & & & & \\
Glove fixed + CRF & 67.9 & 63.4 & 65.6 & 53.7 & 37.6 & 44.2 & 74.1 & 68.1 & 70.9\\
Glove finetuned + CRF & 80.8 & 77.3 & \textbf{79.0} & 63.3 & 45.8 & \textbf{53.1} & 82.1 & 77.0 & \textbf{79.5}\\ \hline \hline
{\em Non-local context only} & & & & & & & & & \\
FW context only + CRF & 71.3 & 39.4 & 50.8 &  53.3 & 19.3 & 28.4 & 71.9 & 58.9 & \textbf{64.7}\\
BW context only + CRF & 69.5 & 47.7 & 56.6 & 46.6 & 21.7 & 29.6 & 74.0 & 49.4 & 59.2\\
BI context only + CRF & 70.1 & 52.1 & \textbf{59.8} & 51.2 & 21.4 & \textbf{30.2} & 66.4 & 56.5 & 61.1\\ \hline \hline
{\em BERT} & & & & & & & & & \\
Full & 91.9 & 93.1 & 92.5 & 75.4 & 75.1 & 75.2 & 96.1 & 97.2 & 96.7 \\ 
Word-only & 80.0 & 80.5 & \textbf{80.3} & 61.6 & 54.8 & \textbf{58.0} & 77.9 & 75.3 & \textbf{76.5}  \\ 
Context-only & 43.1 & 64.1 & 51.6 & 39.7 & 76.2 & 52.2 & 75.6 & 71.6 & 73.5 \\ \hline
\end{tabularx} 
\label{table:Context}
\end{table}

\section{Results}

We evaluate these systems on the CoNLL 2003 and  MUC-6 data. Our goal is to quantify how well the models can work if the identity of the word is not available, and to compare that to situations in which \emph{only} the word identity is known. Additionally, we evaluate the systems trained on CoNLL data on the Wikipedia dataset \cite{balasuriya-etal-2009-named}, to assess how dataset-dependent the performance of the system is. Table \ref{table:Context} reports our results. The first line in the table (BI context + word + CRF) corresponds to the system presented in \citet{huang2015bidirectional}.

\subsection{Word only systems}

The results in the {\em Word only} rows are as expected: We observe low recall and high precision. All systems that rely on the context alone, without taking the identity of the word into account, have worse F1 than the Lookup system. The results are consistent across CoNLL as well as MUC6 datasets. On the cross domain evaluation, however, when the system trained on CoNLL is evaluated on Wikigold, the recall drops considerably.
This behavior may be attributed to the dataset: Many of the entities in the CoNLL training data also appear in testing, a known undesirable fact \cite{augenstein2017generalisation}. 
We find that 64.60\% and 64.08\% of entity tokens in CoNLL and MUC6 test are seen in the respective training sets. However, only 41.55\% of entity tokens in Wikigold are found in the CoNLL training set.

The use of word representations ({\em LogReg} row) contributes substantially to system performance, especially for the MUC-6 dataset in which few names appear in both train and test. Given the impact of the word representations, it would seem important to track how the choice and size of data for training the word representations influences system performance.

\begin{table}
\centering
\small
\setlength{\tabcolsep}{4pt}
\setlength{\extrarowheight}{1pt}
\caption{Mean gate values in CoNLL when entities and non-entities are correct and incorrect. For entities, the average value of context gates remains the same irrespective of the predicted values. For both entities and non-entities, the word gate has a much higher value when the prediction is correct. The word identity itself is the major driver of performance.}
\begin{tabularx}{0.5\linewidth}{L{0.4}|R{0.3}|R{0.3}}
\hline
&\multicolumn{1}{c|}{\bf Context}&\multicolumn{1}{c}{\bf Word}\\\hline
ENT correct & 0.906 & 0.831 \\
ENT incorrect & 0.906 & 0.651 \\ \hline
O correct & 0.613 & 0.897 \\
O incorrect & 0.900 & 0.613 \\\hline
\end{tabularx} 
\label{table:ContextGates}
\end{table}

\subsection{Word and local context combinations}
Next, we consider the systems in the {\em Word + local context} rows. CRFs help to recognize high precision entity-type local contextual constraints, e.g., force a LOC in the pattern `ORG ORG LOC' to be an ORG as well. Another type of high-precision constraining context is word-identity based, similar to the information extraction work discussed above, and constrains X in the pattern `X said' to be PER. Both of these context types were used in  \citet{Liao:2009:SSA:1621829.1621837} for semi-supervised NER.
The observed improved precision and recall of {\em GloVe finetuned + CRF} over {\em LogReg} indicates that the CRF layer modestly improves performance. However, finetuning the representations on the training set is far more important than including such constraints with CRFs as {\em fixed GloVe + CRF} performs consistently worse than {\em LogReg}.

\subsection{Context features only}
We compare {\em Context only} systems with non-local context. In CoNLL data, the context after a word appears to be more predictive, while in MUC-6 the forward context is more predictive. In CoNLL, some of the right contexts are too corpus specific, such as `X 0 Y 1' being predictive of X and Y as organizations with the example occurring in reports of sports games, such as `France 0 Italy 1'. 32.37\% of the {\em ORG} entities in CoNLL test split occur in such sentences. MUC-6, on the other hand, contains many examples that include honorifics, such as `Mr. X'. 35.08\% of PER entities in MUC-6 test split are preceded by an honorific, whereas in CoNLL and MUC6, this is the case only for 2.5\% and 4.6\% entities, respectively. Statistics on these two patterns are shown in Table \ref{table:ContextPatterns}. We provide the list of honorifics and regular expressions for sports scores used to calculate this in the Appendix.

Finally, we note that combining the backward and forward contexts by concatenating their representations results in a better system for CoNLL but not for MUC-6. 

\begin{table}
\centering
\small
\setlength{\tabcolsep}{4pt}
\setlength{\extrarowheight}{1pt}
\caption{Repetitive context patterns in the datasets. In CoNLL, a large percentage of organizations occur in sports scores. In MUC-6, a large percentage of PER entities are preceded by an honorific.}
\begin{tabularx}{0.5\linewidth}{L{0.4}|R{0.3}|R{0.3}}
\hline
&\multicolumn{1}{c|}{\bf Honorifics}&\multicolumn{1}{c}{\bf Sports Scores}\\
& \multicolumn{1}{c|}{\% PER} & \multicolumn{1}{c}{\% ORG} \\\hline
CoNLL train & 2.74 & 25.89 \\
CoNLL testa & 2.19 & 22.13 \\
CoNLL testb & 2.59 & 32.37 \\
Wikipedia & 4.65 & 0.00 \\
MUC-6 train & 27.46 & 0.00 \\
MUC-6 test & 35.08 & 0.00 \\\hline
\end{tabularx} 
\label{table:ContextPatterns}
\end{table}

Clearly, systems with access to only word identity perform better than those with access to only the context (drop of $\sim$20 F1 in all the three datasets). Next, we use the Gated BI + word + CRF system in Figure \ref{fig:gatedsystem} to investigate what the system focuses on when it has access to both the word and to the context, as distinct input representations. We compare the average value of the word and context gates when the system is correct vs incorrect in Table \ref{table:ContextGates}. For entities, while the context gate value is higher than the word gate value, its average remains the same irrespective of whether the prediction is correct or not. On the other hand, the word gate value drops considerably when the system makes an error. Similarly, the word gate value drops considerably for non-entities as well on error. Surprisingly, the context gate value increases for non-entities when an error is made. These results suggest that systems over-rely on word identity to make their predictions. 

Moreover, while one would have expected that the context features have high precision and low recall, this is indeed not the case: the precision of the BI+CRF system is consistently lower than the precision for the full system and the logistic regression model. This means that a better system will not only learn to recognize more contexts but also would be able to override contextual predictions based on features of the word in that context. 

\subsection{Contextualized word representations}
Finally, we experiment with BERT \cite{devlin-etal-2019-bert}. Word-only BERT is finetuned using only one word at a time without any context. Full BERT and context-only BERT are the same system finetuned on the original unchanged dataset and differ only in inference. For context-only BERT, a reserved vocabulary token `[unused0]' is used to hide one word at a time to get a context-only prediction. We report these results in the last two rows of Table \ref{table:Context}. Full BERT improves in F1 over the GloVe based BiLSTM-CRF as reported in the original paper. Word-only BERT performs better than the context-only BERT but the difference in performance is not as pronounced as in the case of the GloVe-based BiLSTM-CRF, except on the CoNLL corpus due to the huge overlap of the training and testing set entities as noted earlier. We also note the difference in performance of the context-only systems using GloVe and BERT. Context-only BERT performs better or worse than context-only BiLSTM, depending on the corpora. These results show BERT is not always better at capturing contextual clues. While it is better in certain cases, it also misses these clues in some instances for which the BiLSTM makes a correct prediction.

In section \ref{sec:Oracle}, we provide a more detailed oracle experiment that demonstrates that the BiLSTM-CRF and BERT context-only systems capture different aspects of context and make errors on different examples. Here we preview this difference in strengths by examining BERT performance on a sample of sentences for which BiLSTM context representations are sufficient to make a correct prediction, and a sample of sentences where it is not. We randomly sampled 200 examples from CoNLL 03 where the context-only LSTM was correct (Sample-C) and another 200 where it was incorrect (Sample-I). Context-only BERT is correct on 71.5\% examples in Sample-C but fails to make the correct prediction on the remaining about 28.5\% that are easy for the BiLSTM context representation. In contrast, it is also able to correctly recognize the entity type in 53.22\% of the cases in Sample-I, where BiLSTM context is insufficient for prediction.

Similar to the gated system, we looked at the attention values in case of BERT. However, these are much harder to interpret as compared to the gate values in the LSTM. The gated LSTM had only one value for the entire context and one value for the entire word. In contrast, BERT has an attention value for each subword in the context and each subword in the word. Moreover, there are multiple attention heads. It is unclear how to interpret the individual values and combine them into a single number for the context and the word. We looked into this by taking the maximum (or mean) value for each subword across all heads and maximum (or mean) value of all subwords in the context and the word but found similar final values for both in all cases.

\begin{table}
\centering
\small
\setlength{\tabcolsep}{3pt}
\setlength{\extrarowheight}{1pt}
\caption{Examples of human evaluation where the context-only system was correct but humans incorrect.}
\begin{tabularx}{\linewidth}{L{0.75}|C{0.13}|C{0.1}|C{0.1}}
\hline
{\bf Sentence} & {\bf Word} & {\bf Label} & {\bf Human}\\\hline
Lang said he \_\_\_ conditions proposed by Britain's Office of Fair Trading, which was asked to examine the case last month. & supported & O & - \\ \hline 
\_\_\_ Vigo 15 5 5 5 17 17 20 & Celta & ORG & O \\ \hline 
The years I spent as manager of the Republic of \_\_\_ were the best years of my life. & Ireland & LOC & - \\ \hline 
\end{tabularx} 
\label{table:HumanSanity}
\end{table}

\section{Human Evaluation}
\label{sec:humaneval}

In this section, we describe a study with humans, assessing if they tend to be more successful at using contextual cues to infer entity types. For comparison, we also conduct another study to assess the success of humans in using just the word to infer entity type.

\begin{table}
\centering
\small
\setlength{\tabcolsep}{2pt}
\setlength{\extrarowheight}{1pt}
\caption{Examples of human evaluation.}
\begin{tabularx}{\linewidth}{L{0.7}|C{0.17}|C{0.12}|C{0.12}|C{0.12}|C{0.12}}
\hline
{\bf Sentence} & {\bf Word} & {\bf Label} & {\bf Human} & {\bf GloVe} & {\bf BERT} \\\hline

\multicolumn{6}{c}{\textbf{Error Class 1}} \\\hline
Analysts said the government , while anxious about \_\_\_ 's debts , is highly unlikely to bring the nickel , copper , cobalt , platinum and platinum group metals producer to its knees or take measures that could significantly affect output . & Norilisk & ORG & \checkmark & O & O  \\ \hline 
6. Germany III ( Dirk Wiese , Jakobs \_\_\_ ) 1:46.02 & Marco & PER & \checkmark & O & O \\ \hline 
- Gulf \_\_\_ Mexico : & of & LOC & \checkmark & MISC & O \\\hline 
About 200 Burmese students marched briefly from troubled Yangon \_\_\_ of Technology in northern Rangoon on Friday towards the University of Yangon six km ( four miles ) away , and returned to their campus , witnesses said . & Institute & ORG & \checkmark & O & LOC \\\hline 

NOTE - Sangetsu Co \_\_\_ is a trader specialising in interiors . & Ltd & ORG & \checkmark & O & \checkmark \\\hline 
Russ Berrie and Co Inc said on Friday that A. \_\_\_ Cooke will retire as president and chief operating officer effective July 1 , 1997 . & Curts & PER & \checkmark & ORG & \checkmark \\\hline 
ASEAN groups Brunei , Indonesia , Malaysia , the \_\_\_ , Singapore , Thailand and Vietnam . & Philippines & LOC & \checkmark & O & \checkmark \\\hline 

\multicolumn{6}{c}{\textbf{Error Class 2}} \\\hline
Their other marksmen were Brazilian defender Vampeta \_\_\_ Belgian striker Luc Nilis , his 14th of the season . & and & O & PER & PER & \checkmark \\\hline 
On Monday and Tuesday , students from the YIT and the university launched street protests against what they called unfair handling by police of a brawl between some of their colleagues and restaurant owners in \_\_\_ . & October & O & LOC & LOC & LOC \\\hline 

Alessandra Mussolini , the granddaughter of \_\_\_ 's Fascist dictator Benito Mussolini , said on Friday she had rejoined the far-right National Alliance ( AN ) party she quit over policy differences last month . & Italy & LOC & PER & O & MISC \\\hline 
Public Service Minister David Dofara , who is the head of the national Red Cross , told Reuters he had seen the bodies of former interior minister \_\_\_ Grelombe and his son , who was not named . & Christophe & PER & ORG & O & \checkmark \\\hline 
The longest wait to load on the West \_\_\_ was 13 days . & Coast & O & MISC & LOC & LOC \\\hline 

, 41 , was put to death in \_\_\_ 's electric chair Friday . & Florida & LOC & - & O & \checkmark \\\hline 
Wall Street , since the bid , has speculated that any deal between Newmont and \_\_\_ Fe would be a "bear hug , " or a reluctantly negotiated agreement where the buyer is not necessarily a friendly suitor . & Santa & ORG & - & LOC & LOC \\\hline 

\end{tabularx} 
\label{table:Human}
\end{table}

\subsection{Context-only}

We performed a human study to determine if humans can infer entity types solely from the context in which they appear. Only sentence-level context is used as all systems operate at the sentence-level. For each instance with a target word in a sentence context, we show three annotators the sentence with the target word masked and ask them to determine its type as PER, ORG, LOC, MISC or O (not a named entity). We allow them to select multiple options. We divide all examples into batches of 10. We ensure quality of the annotation by repeating one example in every batch and removing annotators that are not consistent on this repeated example. Furthermore, we include an example either directly from the instructions or very similar to an example in the instructions. If an annotator does not select the type from the example at a minimum, we assume that they either did not read the instructions carefully, or that they did not understand the task, and we remove them as well. Since humans are allowed to select multiple options, we do not expect them to fully agree on all of the options. The goal is to select the most likely option and so we take as the human answer the option with the most votes: We will refer to this as the majority label.

Similar to the comparison between BiLSTM and BERT, we now compare BiLSTM behavior and that of humans. For the study, we select 20 instances on which the context-only BiLSTM model was correct and 200 instances for which the context-only BiLSTM made errors. The sample from correct prediction is smaller because we see overwhelming evidence that whenever the BiLSTM prediction is correct, humans can also easily infer the correct entity type from contextual features alone. For 85\% of these correctly labeled instances, the majority label provided by the annotators was the same as the true (and predicted) label. 
Table \ref{table:HumanSanity} shows the three (out of 20) examples in which humans did not agree on the category or made a wrong guess. These 20 instances serve as a sanity check as well as a check for annotator quality for the examples where the system made incorrect predictions. 

We received a variety of responses for the 200 instances in sentences where the context only BiLSTM-CRF model made an incorrect prediction. Below we describe the results from the study. We break down the human predictions in two classes. Examples of each are shown in Table \ref{table:Human}.

\paragraph{Error class 1} Humans correct: The human annotators were able to correctly determine the label for 23.5\% (47) of the sentences where the context only BiLSTM-CRF made errors, indicating some room for improvement in a context-only system. 

\paragraph{Error class 2} Human incorrect or no majority: For the remaining 76.5\% instances, humans could not predict the entity type from only the context.

For 55.5\% of the cases, there was a human majority label but it was not the same as the true label. In these cases, the identity of the word would have provided clues conflicting those in the context alone. The large number of such cases --- where context and word cues need to be intelligently combined --- may provide a clue as to why modern NER systems largely ignore context: They do not have the comprehension and reasoning abilities to combine the semantic evidence, and instead resort to over-reliance on the word identity, which in such cases would override the contextual cues in human comprehension.

For 21\%, there was no majority label in the human study, suggesting that the context did not sufficiently constrain the entity type. In such cases, clearly the identity of the words would dominate the interpretation. 

In sum, humans could correctly guess the type without seeing the target word for less than a quarter of the errors made by the BiLSTM-CRF. Remarkably, BERT has correct as well as incorrect predictions on the examples from both BiLSTM-CRF error classes. It was correctly able to determine the entity type for 65.9\% cases in error class 1 and 49.3\% of cases in error class 2. These results show that neither systems is learning the same contextual clues as humans. Humans find the context insufficient in Error class 2 but BERT is able to capture something predictive in the context. Future work could collect more human annotations with humans specifying the reason for selecting an answer. A carefully designed framework would collect human reasoning for their answers and incorporate this information while building a system.

\subsection{Word-only}

For completeness, we also performed human evaluation to determine whether humans can infer the type solely from the word identity. In this case, we do not show the annotators any context. We follow the same annotation and quality control process as above. Because words are ambiguous (Washington can be PER, LOC or ORG), and datasets have different priors for words being of each type, we do not expect the annotators to get all of the answers correct. However, we do expect them to be correct more often than they were in the context-only evaluation.

We select the same 200 instances for the evaluation. We consider the Glove-based LogReg system for comparison as it does not include any context at all. Out of these 200, the LogReg system was correct on 146 (73\%) instances and incorrect for the remaining 27\% instances. Of the 146 cases in LogReg-correct, humans are correct for 116 (79.4\%) instances. For the remaining instances in LogReg-correct, humans are incorrect due to domain/dataset priors. For example, in CoNLL, cities such as Philadelphia are sports teams and hence ORG (not LOC), but without that knowledge, humans categorize them as LOC. Out of all 54 instances in LogReg-incorrect, humans were correct on 16. While this may seem to suggest room for system improvement, it is due to the same ambiguity and dataset bias. For example, CoNLL has a bias for cities like the Philadelphia example above to be sports teams and hence ORGs but there are a few instances where the city name is actually LOC. In these cases, the LogReg system is incorrect but humans are correct.

To summarize, human evaluation using only the word identity is consistent with predictions of the word-only system and observed differences are likely due to dataset biases and so expected.

\section{Oracle Experiments}
\label{sec:Oracle}

In the human evaluation we saw some mixed results, with some room for improvement on 23.5\% of the errors on one side and some errors that seem to be due to over-reliance on context on the other. This motivates investigating whether a more sophisticated approach that decides how to combine cues would be a better approach.

\subsection{Combining all systems}
We perform an oracle experiments where the oracle knows which of the systems is correct. If neither is correct, it defaults to one of the systems. We report results in Table \ref{table:Oracle}. The default system in each case is the one listed first. Row 1 in the table shows that an oracle combination of the forward context only and backward context only does much better than the system which simply concatenates both context representations to make the prediction. The gains are about 15, 20 and 24 points F1 on CoNLL, Wiki and MUC-6 respectively. This improvement captures 22 of the 47 examples (46.8\%) that human annotators got right but not the context-only system in Section \ref{sec:humaneval}.

\begin{table}
\centering
\small
\setlength{\tabcolsep}{3pt}
\setlength{\extrarowheight}{1pt}
\caption{Performance (F1) of GloVe word-based BiLSTM-CRF and BERT. System 1 denotes the oracle combination of separate systems which access to specific input representation only. System 2 refers to a single system with access to all the input of the various systems in system 1. The first two panels were trained on the Original English CoNLL 03 training data and tested on the original English CoNLL 03 test data and the WikiGold data. The last panel was trained and tested on the respective splits of MUC-6.}
\begin{tabularx}{\linewidth}{L{0.7}|L{0.35}|C{0.21}C{0.21}|C{0.21}C{0.21}|C{0.21}C{0.21}}
\hline
\multirow{2}{*}{\textbf{System 1 (Oracle)}}& \multirow{2}{*}{\textbf{System 2}}&\multicolumn{2}{c|}{\textbf{CoNLL}} & \multicolumn{2}{c|}{\textbf{Wikipedia}}& \multicolumn{2}{c}{\textbf{MUC-6}} \\
\cline{3-8}
& & sys 1 & sys 2 & sys 1 & sys 2 & sys 1 & sys 2\\\hline
\rule{-2pt}{8pt}
FW context -- BW context LSTM-CRF & Bi context LSTM-CRF & 75.3 & 59.8 & 49.3 & 30.2 & 85.3 & 61.1 \\ \hline
\rule{-2pt}{18pt}
\makecell[l]{FW context -- BW context\\ -- Glove finetuned \\LSTM-CRF} & Full LSTM-CRF & 92.2 & 91.0 & 72.4 & 63.6 & 94.9 & 90.9\\ \hline
\rule{-2pt}{18pt}
\makecell[l]{Full system -- FW context\\ -- BW context -- Glove\\ finetuned LSTM-CRF} & Full LSTM-CRF & 95.1 & 91.0 & 76.8 & 63.6 & 96.1 & 90.9\\ \hline
\rule{-2pt}{12pt}
\makecell[l]{word-only -- context-only \\ BERT} & Full BERT & 92.5 & 92.5 & 86.8 & 75.2 & 93.7 & 96.7\\ \hline
\rule{-2pt}{18pt}
\makecell[l]{Full -- word-only \\-- context-only BERT} & Full BERT & 96.5 & 92.5 & 90.1 & 75.2 & 98.5 & 96.7 \\ \hline
\end{tabularx} 
\label{table:Oracle}
\end{table}

\begin{table}
\centering
\small
\setlength{\tabcolsep}{3pt}
\setlength{\extrarowheight}{1pt}
\caption{Oracle combination of the the context-only systems using GloVe and BERT. {\em min} and {\em max} denote the minimum and maximum F1 of the systems combined and {\em comb} denotes the F1 of the combined system. The first two panels were trained on the Original English CoNLL 03 training data and tested on the original English CoNLL 03 test data and the WikiGold data. The last panel was trained and tested on the respective splits of MUC-6.}
\begin{tabularx}{\linewidth}{L{0.85}|C{0.2}C{0.2}C{0.2}|C{0.2}C{0.2}C{0.2}|C{0.2}C{0.2}C{0.2}}
\hline
\multirow{2}{*}{\textbf{System}}& \multicolumn{3}{c|}{\textbf{CoNLL}} & \multicolumn{3}{c|}{\textbf{Wikipedia}}& \multicolumn{3}{c}{\textbf{MUC-6}} \\
\cline{2-10}
&min&max&comb&min&max&comb&min&max&comb\\\hline
\rule{-2pt}{8pt}
\makecell[l]{context-only LSTM -- \\ context-only BERT} & 51.6 & 59.8 & 83.7 & 30.2 & 52.2 & 79.0 & 61.1 & 73.5 & 84.6  \\ \hline
\end{tabularx} 
\label{table:Oracle_Encodings}
\end{table}

We performed more such experiments with the full system and the word-only and context-only systems. These are shown in row 2 and 3. In each case, there are gains over the full BiLSTM-CRF. An oracle with the four systems (row 3) shows the highest gains with $\sim$4 points F1 on CoNLL and 6 points F1 on MUC-6. The gains are especially pronounced in case of cross-domain evaluation i.e. the system trained on CoNLL when evaluated on Wikipedia has an increase of 13 points F1. 

These results indicate that when given access to different components -- word, forward context, backward context -- systems recognize different entities correctly, as they should. However, when all of these components are thrown at the system at once, they are not able to combine these in the best possible way.

\subsection{Combining BERT variants}
We performed similar experiments with the full, word-only and context-only BERT as well (row 4 and 5). Remember that the context-only BERT is the same trained system as the full BERT and the difference comes from masking the word during evaluation. The word-only and context-only combination shows no improvement over full BERT on CoNLL but does so on the two other datasets. However, even for CoNLL, the oracle of these two systems is correct on somewhat different examples than the full BERT as evident in the last row where combining all three systems gives improvement on all three datasets. Again, the improvement is highest on cross-domain evaluation on the Wikipedia dataset as with the combination of the GloVe based systems. We hypothesize that the attention on the current word likely dominates in the output contextual representation, aggravated even further because of softmax because of which the relevant context words do not end up contributing much. However, when the word is masked in the context-only variant, the attention weight on it will likely be small because we use one of the tokens unused in the vocabulary and the relevant context words end up contributing to the final representation. Future work could involve some regularization to reduce the impact of the current word in few random training examples so that systems do not overfit to the word identity and focus more on the context. 

\subsection{Combining context-only LSTM and BERT}
Lastly, we performed an oracle experiment with the context-only LSTM-CRF and the context-only BERT. We see the biggest jumps in performance with this particular experiment -- 24, 27 and 11 points F1 on CoNLL, Wiki and MUC-6 respectively. Both the systems use the context differently and are correct on a large number of different examples, again showing that better utilization of context is possible. Note that the human study showed that the room for improvement from context is small. The study was based on 200 random examples, where BERT was able to get some correct even when humans could not. We hypothesize that while the context was possibly not sufficient to identify the type, a similar context had been seen in the training data which was learned by BERT.

All the oracle experiments show room for improvement and future work would involve looking into strategies/systems to combine these components better. The progress towards this can be measured by breaking down the systems and conducting oracle experiments as here.

\section{Discussion and future Work}
\label{sec:future}

We started the paper with three questions about the workings of named entity recognition systems. Here, we return to these questions and synthesize the insights related to each gained from the experiments presented in the paper. We zeroed in on the question of interpretability of named entity recognition systems, specifically examining the performance of systems that represent differently the current word, the context and their combination. We tested the systems on two corpora and one tested across domains and show that some of the answers to these questions are times corpus dependent.

\subsection{What Do Systems Learn?} 
 {\em Word types, mostly}. 
 
 We find that current systems, including those build on top of contextualized word representations, pay more attention to the current word than to the contextual features. Partly this is due to the fact that contextual features do not have high precision and in many cases need to be overridden by evidence from the current word. Moreover, we find that contextual representations, namely BERT are not always better at capturing context as compared to systems such as Glove-based BiLSTM-CRFs. Their higher performance could be results of better pretraining data and learning the subword structure better. We leave this analysis for future work and instead focus on the extent of word vs context utilization with more focus on context utilization for better generalization.

\subsection{Can Context Be Utilized Better?}
{\em Only a bit better, if we want to emulate humans. But a lot better if willing to incorporate the superhuman abilities of transformer models.} 

We carry out a study to test the ability of human annotators to predict the type of an entity without seeing the entity word. Humans seem to easily do the task on examples where the context-only system predicts the entity type correctly. The examples on which the context-only system makes a mistake are difficult for humans as well. Humans can guess the correct label only in about a quarter of all such examples. The opportunity for improvement from better contextual representations that recognize more constraining contexts exists but is relatively small. 

\subsection{How Can We Utilize Context Better?}

{\em By developing reasoning, more flexible combination of evidence and informed regularization.} Based on our experiments, several avenues for better context utilization emerge.

\paragraph{Human-like reasoning} Our human study shows that systems are not capturing the same information from textual context as humans do. BERT is able in many cases to correctly recognize the type from the context even when humans fail to do so. A direction for future work would involve collecting human reasoning behind their answers and incorporating this information in building the systems. This means learning to identify when the context is constraining and possibly what makes it constraining.
    
\paragraph{Avoiding concatenation} Another promising direction for the overall improvement of GloVe-based BiLSTM-CRF system appears to be the better combination of features representing different types of context and the word. Oracle experiments show that different parts of the sentence -- word, forward context, backward context -- can help recognize entities correctly when used standalone but not when used together. A simple concatenation of features is not as meaningful and that a smarter combination of several types of features can lead to better performance.
    
\paragraph{Attention regularization} Oracle experiments involving BERT show that hiding the word itself can sometimes correctly identify the word type even when seeing the word leads to an incorrect prediction. BERT uses attention weights to combine different parts of the input instead of concatenation as in the just discussed BiLSTM approaches. We hypothesize that the attention on the word is likely much larger than on the rest of the input because of which the relevant context words do not end up contributing much. Future work could involve some regularization to reduce the impact of the current word in few random training examples so that it does not overfit to the word identity and can focus more on the context. 

Lastly, another direction for future work could expand the vocabulary of entities instead of trying to learn context better so that more entities are seen either directly in the training data or have similar representation in the embedding space by virtue of being seen in the pretraining data. This could be done by having an even larger pre-training data from diverse sources for better coverage or by incorporating resources such as knowledge bases and gazetteers in the contextual systems. 

\section*{Acknowledgements}

This material is based upon work supported in part by the National Science Foundation under Grant No. (NSF 1901117).
\starttwocolumn
\bibliography{compling_style}

\appendix

\section{Entity Recognition vs Typing}
Entity recognition results are shown in Table \ref{table:EntityRecognition}. Here, we check if a word is correctly recognized as an entity, even if the type is wrong. The results are better than Recognition+Typing results in Table \ref{table:Context} for all systems and datasets but still not very high. Therefore, the errors made are in both recognition and typing. The same performance pattern is observed with word only systems better than context-only systems even in entity recognition. The word-only systems have an F1 much higher than context-only systems, meaning that looking at a word, it is much easier to say whether it is an entity or not. But looking at context, it is still hard to say whether something is an entity or not. The breakdown by type is also shown in Table \ref{table:Confusion matrix} for all three datasets for Full BERT.

\begin{table*}
\centering
\small
\setlength{\tabcolsep}{3pt}
\setlength{\extrarowheight}{1pt}
\caption{Performance of GloVe word-level BiLSTM-CRF and BERT on entity recognition (no typing). All rows are for the former and only the last two rows for BERT. Local context refers to high precision constraints due to sequential CRF. Non-local context refers to the entire sentence. No document level context is included. The first two panels were trained on the Original English CoNLL 03 training data and tested on the original English CoNLL 03 test data and the WikiGold data. The last panel was trained and tested on the respective splits of MUC-6.}
\begin{tabularx}{\linewidth}{L{0.8}|C{0.2}C{0.2}C{0.2}|C{0.2}C{0.2}C{0.2}|C{0.2}C{0.2}C{0.2}}
\hline
\multirow{2}{*}{\textbf{System}}& \multicolumn{3}{c|}{\textbf{CoNLL}} & \multicolumn{3}{c|}{\textbf{Wikipedia}}& \multicolumn{3}{c}{\textbf{MUC-6}} \\
\cline{2-10}
&P&R&F1&P&R&F1&P&R&F1\\\hline
{\em Full system} & & & & & & & & & \\
BI context + word + CRF & 96.2 & 96.8 & 96.5 & 86.7 & 79.2 & 82.8 & 94.5 & 96.3 & 95.4 \\  \hline \hline
{\em Words only} & & & & & & & & & \\
Lookup & 96.2 & 64.8 & 77.4 & 93.4 & 40.1 & 56.1 & 93.6 & 62.6 & 75.0 \\
LogReg & 93.8 & 86.9 & 90.2 & 89.9 & 74.8 & 81.6 & 87.5 & 83.6 & 85.5\\ \hline \hline
{\em Words + local context} & & & & & & & & & \\
Glove fixed + CRF & 83.2 & 77.6 & 80.3 & 81.6 & 57.1 & 67.2 & 83.5 & 76.7 & 80.0 \\
Glove finetuned + CRF & 91.9 & 88.0 & 89.9 & 88.6 & 64.1 & 74.4 & 91.3 & 85.6 & 88.4 \\ \hline \hline
{\em Non-local context only} & & & & & & & & & \\
FW context only + CRF & 78.8 & 43.6 & 56.1 & 65.8 & 23.9 & 35.0 & 76.6 & 62.8 & 69.0 \\
BW context only + CRF & 76.6 & 52.6 & 62.3 & 62.0 & 28.9 & 39.5 & 76.8 & 51.2 & 61.4 \\
BI context only + CRF & 78.0 & 58.0 & 66.5 & 64.8 & 27.1 & 38.2 & 72.1 & 61.4 & 66.3 \\ \hline \hline
{\em BERT} & & & & & & & & & \\
Full & 97.0 & 98.3 & 97.7 & 88.3 & 87.9 & 88.1 & 97.2 & 98.3 & 97.7 \\ 
Word-only & 94.8 & 95.4 & 95.1 & 93.0 & 82.8 & 87.6 & 93.1 & 89.9 & 91.5 \\ 
Context-only & 50.3 & 74.7 & 60.1 & 47.7 & 91.5 & 62.7 & 77.7 & 73.5 & 75.5 \\ \hline
\end{tabularx} 
\label{table:EntityRecognition}
\end{table*}

\begin{table}
\centering
\small
\setlength{\tabcolsep}{3pt}
\setlength{\extrarowheight}{1pt}
\caption{Confusion matrices for BERT-Full. Rows represent the true class and columns represent the predicted class.}
\begin{tabularx}{\linewidth}{L{0.15}|R{0.2}R{0.2}R{0.2}R{0.2}R{0.2}}
\hline
& \multicolumn{1}{c}{PER} & \multicolumn{1}{c}{ORG} & \multicolumn{1}{c}{LOC} & \multicolumn{1}{c}{MISC} & \multicolumn{1}{c}{O} \\ 
\hline
\multicolumn{1}{c}{\bf CoNLL}\\
\hline
PER & 2709 & 33 & 15 & 5 & 11 \\
ORG & 8 & 2320 & 86 & 47 & 35 \\
LOC & 6 & 72 & 1789 & 39 & 19 \\
MISC & 10 & 77 & 25 & 736 & 70 \\
O & 18 & 81 & 22 & 126 & 38076 \\ 
\hline
\multicolumn{1}{c}{\bf Wikipedia} \\
\hline
PER & 1509 & 14 & 13 & 6 & 92 \\
ORG & 50 & 1362 & 200 & 192 & 154 \\
LOC & 14 & 141 & 1013 & 39 & 240 \\
MISC & 22 & 74 & 61 & 944 & 291 \\
O & 51 & 154 & 230 & 312 & 31827 \\
\hline
\multicolumn{1}{c}{\bf MUC-6} \\
\hline
PER & 584 & 4 & 1 & 0 & 1 \\
ORG & 6 & 830 & 5 & 0 & 22 \\
LOC & 0 & 0 & 105 & 0 & 4 \\
MISC & 0 & 0 & 0 & 0 & 0 \\
O & 3 & 41 & 1 & 0 & 12498 \\
\hline
\end{tabularx} 
\label{table:Confusion matrix}
\end{table}

\section{Honorifics}
Datasets have different common context patterns. Example, forward context is highly predictive in MUC-6. 35.08\% of PER entities in MUC-6 are preceded by an honorific whereas in CoNLL and MUC6, this number is only 2.5\% and 4.6\% respectively. Following is the list of honorifics (mostly English, owing to the nature of the datasets) used to calculate these numbers - Dr, Mr, Ms, Mrs, Mstr, Miss, Dr., Mr., Ms., Mrs., Mx., Mstr., Mister, Professor, Doctor, President, Senator, Judge, Governor, Officer, General, Nurse, Captain, Coach, Reverend, Rabbi, Ma'am, Sir, Father, Maestro, Madam, Colonel, Gentleman, Sire, Mistress, Lord, Lady, Esq, Excellency, Chancellor, Warden, Principal, Provost, Headmaster, Headmistress, Director, Regent, Dean, Chairman, Chairwoman, Chairperson, Pastor.

\section{Sports Scores}
We use the following three regular expressions to determine if a sentence contains sports scores - 

\noindent\centerline{\scriptsize ([0-9]+. )?([A-Za-z]+ )\{1,3\}([0-9]+ )\{0,6\}(([0-9]+)(?!\/))( 1\/2)?( -)?}  \\
\centerline{\scriptsize and} \\
\centerline{\scriptsize ([A-Za-z]+ )\{1,3\}([0-9]+ )\{1,3\}([A-Za-z]+ )\{1,3\}([0-9]+ )\{0,2\}[0-9]} \\
\centerline{\scriptsize and} \\
\centerline{\scriptsize ([A-Z]+ )\{1,3\}AT ([A-Z]+ )\{1,2\}[A-Z]+}

\begin{table*}
\centering
\footnotesize
\setlength{\tabcolsep}{2pt}
\caption{Breakdown of human labels. The majority label is selected as the final label.}
\begin{tabularx}{\linewidth}{L{1.2}|C{0.23}|C{0.15}|C{0.1}C{0.1}C{0.1}C{0.1}C{0.1}}
\hline
\multirow{2}{*}{\bf Sentence} & \multirow{2}{*}{\bf Word} & {\bf True} & \multicolumn{4}{c}{\bf Human label} \\
& & {\bf Label} & {\bf PER} & {\bf ORG} & {\bf LOC} & {\bf MISC} & {\bf O} \\\hline

\multicolumn{6}{c}{\textbf{Error Class 1}} \\\hline
Analysts said the government , while anxious about \_\_\_ 's debts , is highly unlikely to bring the nickel , copper , cobalt , platinum and platinum group metals producer to its knees or take measures that could significantly affect output . & Norilisk & ORG & 0 & 2 & 1 & 0 & 0 \\ \hline 
6. Germany III ( Dirk Wiese , Jakobs \_\_\_ ) 1:46.02 & Marco & PER & 2 & 1 & 1 & 1 & 0 \\ \hline 
- Gulf \_\_\_ Mexico : & of & LOC & 1 & 0 & 2 & 0 & 0 \\\hline 
About 200 Burmese students marched briefly from troubled Yangon \_\_\_ of Technology in northern Rangoon on Friday towards the University of Yangon six km ( four miles ) away , and returned to their campus , witnesses said . & Institute & ORG & 0 & 2 & 1 & 1 & 0 \\\hline 

NOTE - Sangetsu Co \_\_\_ is a trader specialising in interiors . & Ltd & ORG & 0 & 3 & 0 & 0 & 0 \\\hline 
Russ Berrie and Co Inc said on Friday that A. \_\_\_ Cooke will retire as president and chief operating officer effective July 1 , 1997 . & Curts & PER & 3 & 0 & 0 & 0 & 0 \\\hline 
ASEAN groups Brunei , Indonesia , Malaysia , the \_\_\_ , Singapore , Thailand and Vietnam . & Philippines & LOC & 1 & 0 & 2 & 0 & 0 \\\hline 

\multicolumn{6}{c}{\textbf{Error Class 2}} \\\hline
Their other marksmen were Brazilian defender Vampeta \_\_\_ Belgian striker Luc Nilis , his 14th of the season . & and & O & 2 & 0 & 1 & 0 & 0 \\\hline 
On Monday and Tuesday , students from the YIT and the university launched street protests against what they called unfair handling by police of a brawl between some of their colleagues and restaurant owners in \_\_\_ . & October & O & 1 & 0 & 3 & 0 & 0 \\\hline 

Alessandra Mussolini , the granddaughter of \_\_\_ 's Fascist dictator Benito Mussolini , said on Friday she had rejoined the far-right National Alliance ( AN ) party she quit over policy differences last month . & Italy & LOC & 3 & 0 & 1 & 1 & 0 \\\hline 
Public Service Minister David Dofara , who is the head of the national Red Cross , told Reuters he had seen the bodies of former interior minister \_\_\_ Grelombe and his son , who was not named . & Christophe & PER & 1 & 2 & 0 & 0 & 0 \\\hline 
The longest wait to load on the West \_\_\_ was 13 days . & Coast & O & 1 & 0 & 0 & 2 & 0 \\\hline 

, 41 , was put to death in \_\_\_ 's electric chair Friday . & Florida & LOC & 0 & 2 & 2 & 0 & 0 \\\hline 
Wall Street , since the bid , has speculated that any deal between Newmont and \_\_\_ Fe would be a "bear hug , " or a reluctantly negotiated agreement where the buyer is not necessarily a friendly suitor . & Santa & ORG & 2 & 2 & 1 & 0 & 0 \\\hline 

\multicolumn{6}{c}{\textbf{Sanity check errors}} \\\hline
Lang said he \_\_\_ conditions proposed by Britain's Office of Fair Trading, which was asked to examine the case last month. & supported & O & 0 & 0 & 0 & 2 & 2 \\ \hline 
\_\_\_ Vigo 15 5 5 5 17 17 20 & Celta & ORG & 0 & 1 & 0 & 0 & 2 \\ \hline 
The years I spent as manager of the Republic of \_\_\_ were the best years of my life. & Ireland & LOC & 1 & 2 & 2 & 0 & 0 \\ \hline 

\end{tabularx} 
\label{table:HumanBreakdown}
\end{table*}

\section{Human Labels Breakdown}
Here we provide the breakdown between the different labels as marked by the human annotators. Note that the annotators could select multiple options and the majority label was selected as the final label. The results are shown in Table \ref{table:HumanBreakdown}.

\end{document}